\documentclass{article}
 
\usepackage[eandd, final]{neurips_2026_my}
 
\usepackage[utf8]{inputenc}
\usepackage[T1]{fontenc}
\usepackage{hyperref}
\usepackage{url}
\usepackage{booktabs}
\usepackage{amsfonts}
\usepackage{nicefrac}
\usepackage{microtype}
\usepackage{xcolor}
\usepackage{subfiles}
\usepackage{array}
\usepackage{multirow}
\usepackage{graphicx}
\usepackage{amssymb}
\usepackage{amsmath}
\usepackage{makecell}
\usepackage{rotating}
\usepackage{float}
\usepackage{caption}
\usepackage{subcaption}
\usepackage{enumitem}
 
\hypersetup{
    colorlinks=true,
    linkcolor=blue,
    citecolor=blue,
    urlcolor=blue
}
 
\newcommand{\Paragraph}[1]{\vspace{0mm} \noindent \textbf{#1} \hspace{0mm}}
 
\title{What Do Deepfake Benchmarks Measure? \\
An Audit Using Frozen Self-Supervised Representations}

\author{\mdseries%
  Samuel Pagon$^{1}$ \quad          %
  Yixuan Shen$^{1}$ \quad        %
  Vishal Asnani$^{2}$ \quad
  Feng Liu$^{1}$ \quad           %
  \\
  $^1$ Department of Computer Science, Drexel University\\
  $^2$ Adobe Research\\
  \texttt{vasnani@adobe.com, \{sp3692,ys844,fl397\}@drexel.edu} \\
}

\begin{document}

\maketitle

\begin{abstract}
As deepfake generators approach perceptual indistinguishability, reliable detection becomes critical. Yet, detectors that score well on benchmarks routinely fail in the wild. A concerning feedback loop has emerged: benchmarks drive increasingly complex, engineered detectors, yet if those benchmarks do not reflect real-world deepfakes, this complexity may be solving the wrong problem entirely. This raises a prior question: what are these benchmarks actually measuring? We conduct an audit of video, image, and audio deepfake benchmarks using a deliberately simple diagnostic. If a linear probe on frozen, general-purpose self-supervised representations can approximate the performance of a bespoke detector, the benchmark is largely rewarding general modality understanding rather than forensic understanding. This has two implications: the benchmark may not reflect realistic threat models, and it raises the question of whether the bespoke detectors the probe approaches are truly learning forensic understanding. We observe, across three modalities, linear probes on general-purpose self-supervised representations closely approach the performance of bespoke detectors. We further show that generator-level difficulty is partly explained by Fréchet geometry in the same representation space. Together, these results support a benchmark-audit view of deepfake detection: before high scores are read as evidence of forensic understanding, it is worth asking how much of the benchmark is already solved by general-purpose representations.
\end{abstract}

\section{Introduction}\label{sec:intro}
Deepfakes are now a routine vector for fraud and political manipulation. Recent 2026 incidents include synthetic videos impersonating the Governor of the Bank of Italy in investment scams~\cite{reuters2026panetta}, deepfake celebrity endorsements driving crypto schemes flagged by state attorneys general~\cite{agny2026meta,ncdoj2026meta}, and AI-generated political ads in the U.S. midterm cycle~\cite{reuters2026midtermdeepfakes}. Reliable detection is what current deepfake benchmarks are meant to produce, but the gap between benchmark and deployed performance is well-documented. \emph{Deepfake-Eval-2024}~\cite{chandra2025deepfake} and the \emph{SAFE Challenge}~\cite{trapeznikov2025synthetic} show that detectors strong on standard splits collapse on in-the-wild content, and recent cross-benchmark analysis attributes much of the apparent progress to shortcut learning rather than forensic understanding~\cite{yermakov2026deepfake}. Benchmark performance keeps climbing, but generalization to the deepfakes that actually matter does not follow. This raises a prior question: \emph{\textbf{what are these benchmarks actually measuring?}}

To find out we audit the benchmarks themselves rather than propose another detector. For each modality we fit a linear classifier on layer-wise representations from a frozen pretrained self-supervised model, trained for general modality understanding. This ensures the probe can only exploit structure that a general-purpose representation of the modality already encodes. Its performance then reveals how much of a benchmark is solvable without any forensic understanding. We complement the probe with a representation-space geometry analysis that asks why some target generators are harder than others, using each generator's position relative to source-real and source-fake distributions in the same frozen representation space.

Our contributions are:
\begin{itemize}[
    leftmargin=1.2em,
    labelsep=0.5em,
    itemsep=1pt,
    topsep=2pt,
    parsep=0pt,
    partopsep=0pt
]
    \item A benchmark-audit framework applicable across image, audio, and video, built around pretrained frozen self-supervised models and linear probes.
    \item Empirical evidence that across three contemporary generator-diverse benchmarks (AIGVDBench, Celeb-DF++, MLAAD v9 English, validated on ASVspoof2019 LA), frozen-SSL probes approach the performance of specialized detectors. Showing substantial real-versus-fake signal is already linearly accessible in representation spaces never learned for deepfake detection.
    \item A geometric account of generator-level difficulty showing that the relative Fr\'echet margin between source-real and source-fake regions is the most consistent predictor of which target generators are easy or hard for the probe, surfacing a structural concern that some generators in current test sets sit close to already-seen spoof structure rather than testing genuinely novel generators.
\end{itemize}

\section{Related Work}\label{sec:related}
\Paragraph{Deepfake Benchmarks and the Generalization Gap.}
Deepfake benchmarks have evolved along two axes: broader scope and more transfer-oriented evaluation. Early face benchmarks like \emph{FaceForensics++} gave way to larger, more realistic datasets such as \emph{Celeb-DF}, \emph{DFDC}, and \emph{DeeperForensics-1.0}~\cite{rossler2019faceforensics++,li2020celebdf,dolhansky2020deepfake,jiang2020deeperforensics}, while in speech \emph{ASVspoof 2019} established the canonical logical-access anti-spoofing protocol~\cite{nautsch2021asvspoof}. Recent resources emphasize generator coverage at contemporary scale, spanning standardized cross-dataset protocols (\emph{DeepfakeBench}), multilingual and multi-generator audio (\emph{MLAAD}), expanded face-video scenarios across face-swap, face-reenactment, and talking-face (\emph{Celeb-DF++}), and large-scale AI-video suites (\emph{GenVidBench}, \emph{AIGVDBench})~\cite{yan2023deepfakebench,muller2024mlaad,li2025celeb,ni2026genvidbench,ma2026your}. A recurring theme is that strong in-distribution performance fails to translate into real-world robustness: \emph{Deepfake-Eval-2024} reports large AUC drops for open-source detectors on in-the-wild content~\cite{chandra2025deepfake}, and the \emph{SAFE} Challenge shows audio detectors degrade sharply under post-processing and laundering~\cite{trapeznikov2025synthetic}.

\Paragraph{Self-supervised representations in deepfake detection.}
Self-supervised and foundation-model representations are increasingly central to deepfake detection. \emph{RealForensics} demonstrated that SSL pretraining on real talking faces improves robustness to unseen manipulations~\cite{haliassos2022leveraging}; \emph{UnivFD} showed that simple nearest-neighbor and linear-probe baselines on CLIP features generalize well to unseen diffusion and autoregressive models~\cite{ojha2023towards}; subsequent work reinforces this finding for CLIP-based and SSL-pretrained ViT detectors~\cite{cozzolino2024raising,nguyen2024exploring,cui2025forensics}. The same pattern holds in audio: \emph{Spoof-SUPERB} evaluates twenty SSL speech models for anti-spoofing and finds that large discriminative encoders such as XLS-R, UniSpeech-SAT, and WavLM Large transfer best~\cite{ali2026superb}. Across modalities, generic pretrained representation spaces already contain substantial fake/real signal, as defined by the benchmarks. Our work builds on this insight by using frozen SSL representations not primarily as better detectors, but as \emph{benchmark diagnostics}.

\Paragraph{Distributional geometry and benchmark analysis.}
Fr\'echet-style distances motivate representation-space distribution comparison. \emph{Fr\'echet Inception Distance} (FID) compares Gaussian approximations of real and generated image embeddings~\cite{heusel2017gans}, with related variants \emph{FAD}~\cite{kilgour2018fr} and \emph{FVD}~\cite{unterthiner2018towards} for audio and video, and nonparametric alternatives such as \emph{MMD}~\cite{gretton2012kernel}. Recent work also clarifies the limits of geometry-based evaluation: \emph{On the Content Bias in Fr\'echet Video Distance} shows that FVD can be biased toward per-frame quality and that SSL video representations reduce this bias~\cite{ge2024content}. In deepfake detection, recent cross-benchmark analysis points to shortcut learning and training-set composition as major drivers of apparent generalization, rather than simple chronological benchmark difficulty~\cite{yermakov2026deepfake}. Together, these works motivate our use of SSL representation-space geometry as a benchmark-audit tool: if generator difficulty is predictable from where a target generator distribution lies relative to source-real and source-fake regions, then benchmark scores partly reflect representation-space coverage rather than detector-specific forensic understanding.

\section{Audit Protocol}\label{sec:method}

Our audit framework has two stages applied uniformly across modalities. First, we use a frozen transformer-based SSL to convert each input into layer-wise pooled representations. Second, we train simple linear probes on those representations and analyze the geometry of the same frozen SSL representation space. By keeping the backbone fixed and the head linear, we test whether strong benchmark performance is already accessible in generic pretrained representation spaces, rather than arising from task-specific detector engineering.

\subsection{Benchmarks}

We evaluate three benchmarks, each chosen due to their advertised coverage of generators, spanning from legacy to recent state-of-the-art image, audio, and video generation quality. Table~\ref{tab:benchmarks} summarizes the protocols. In each case the source domain is held fixed for training, and the target domain is held fixed for evaluation, so that any reported gap reflects generator and dataset transfer rather than within-dataset memorization.

\begin{table}[h]
\centering
\caption{Source--target benchmark pairs used in the audit. Sample counts and protocol details follow the original benchmark releases. MLAAD evaluation is restricted to the English split because ASVspoof2019 LA is English-only; this isolates generator shift from cross-lingual mismatch.}
\label{tab:benchmarks}
\small
\setlength{\tabcolsep}{4pt}
\renewcommand{\arraystretch}{1.15}
\begin{tabular}{p{1.0cm} p{3.0cm} p{1.6cm} p{5.5cm} p{1.7cm}}
\toprule
\textbf{Modality} & \textbf{Source $\rightarrow$ Target} & \textbf{Backbone} & \textbf{Key protocol details} & \textbf{Metric} \\
\midrule
Audio &
ASVspoof2019 LA $\rightarrow$ MLAAD v9 (English) &
\texttt{XLS-R 300M~\cite{babu2021xls}} &
ASVspoof2019 LA: 2{,}580 / 22{,}800 (bona fide / spoof) train, 2{,}548 / 22{,}296 dev, 7{,}355 / 63{,}882 eval. MLAAD v9: 687.4 hrs, 140 TTS systems, 51 languages; English split has 84 generators in our experiments. &
EER (LA), spoof-only generator accuracy (MLAAD v9) \\
\midrule
Image &
Celeb-DF v2 $\rightarrow$ Celeb-DF++ &
\texttt{DINOv3 ViT-L~\cite{simeoni2025dinov3}} &
Celeb-DF v2: 590 real, 5{,}639 fake, 59 celebrities, $>$2M frames. Celeb-DF++: 53{,}196 fakes from 22 methods across face-swap, face-reenactment, talking-face. Target test split uses 178 reals; 200 fakes per FS method, 200 per FR method, 300 per TF method. &
Per-generator AUC vs.~shared real pool \\
\midrule
Video &
AIGVDBench &
\texttt{V-JEPA2 ViT-G~\cite{assran2025v}} &
AIGVDBench: $>$440{,}000 videos from 31 methods (20 open-source, 11 closed-source). Open-source: 14k/3k/3k train/val/test per generator with matched reals. Closed-source: 2{,}000 test videos per generator. Training uses real videos plus \texttt{Open-Sora} fakes; evaluation uses the full generator-diverse test split. &
Per-generator AUC vs.~shared real pool \\
\bottomrule
\end{tabular}
\end{table}

To make the source--target splits visually inspectable, we also provide an interactive benchmark viewer in Appendix, 
which samples real and fake examples from the training splits and one example per target generator across video, image, and audio.

\subsection{Layer-wise representation extraction and linear probes}

Let $f_{\theta}$ denote a pretrained transformer encoder with parameters $\theta$ and let $\ell \in \{1,\dots,L\}$ index its hidden layers. For an input $x_i$, the encoder returns a sequence of hidden tokens
\(
H_i^{(\ell)} = [h_{i,1}^{(\ell)},\dots,h_{i,T_\ell}^{(\ell)}] \in \mathbb{R}^{T_\ell \times d_\ell},
\)
where $T_\ell$ is the number of tokens and $d_\ell$ the hidden dimensionality. Tokens correspond to time steps for audio and to image patches or video tokens for the visual modalities. We obtain a fixed-dimensional vector by global mean pooling:
\begin{equation}
    z_i^{(\ell)} = \frac{1}{T_\ell}\sum_{t=1}^{T_\ell} h_{i,t}^{(\ell)} \in \mathbb{R}^{d_\ell}.
\end{equation}
Each layer's source-domain training representations are then standardized via z-score normalization, and the same statistics are applied to target representations.

\Paragraph{Probes.}
We fit two linear probes independently at every layer: $L_2$-regularized Logistic Regression and an $L_2$-regularized Ridge classifier. For AUC and EER computation, we use the logistic probability for Logistic Regression and the raw linear decision score for Ridge. Both probes are deliberately low-capacity, restricting them from learning new deep representations. If such probes achieve performance close to specialized detectors, the benchmark is at least partly solvable through linearly accessible structure already present in the frozen SSL representation space.

\Paragraph{Modality-specific implementation.}
Audio waveforms are downmixed to mono, resampled to 16 kHz, and trimmed or zero-padded to 64{,}000 samples; MLAAD items are assigned the spoof label, so MLAAD performance is reported as per-generator spoof recall. Image representations are extracted from aligned face crops at $224{\times}224$: Celeb-DF v2 uses keyframes, Celeb-DF++ uses 32 deterministically sampled frames per video, and both pipelines run a YuNet-based~\cite{wu2023yunet} eye-keypoint alignment. Video representations use a fixed sequence of 8 frames per clip (the final 8 frames if available, cycled otherwise). Probes are fit on source-domain labels and evaluated on the target test split. For the video benchmark, source training is restricted to real videos plus \texttt{Open-Sora}~\cite{opensora} fakes; all other generators are excluded from the source split.

\subsection{Representation-space geometry}

To analyze why some target generators are easier than others, we model each generator group as a Gaussian in SSL representation space. For group $g$ at layer $\ell$, with members $G_g$,
\begin{equation}
    \mu_g^{(\ell)} = \frac{1}{|G_g|}\sum_{i\in G_g} z_i^{(\ell)}, \qquad
    \Sigma_g^{(\ell)} = \mathrm{Cov}\!\left(\{z_i^{(\ell)}\}_{i\in G_g}\right).
\end{equation}
Group definitions are benchmark-specific. For audio, the source domain has one real group and multiple spoof groups (with no target real group, since MLAAD is spoof-only); for image, the source domain has one real and one aggregated fake group, while the target domain has reals plus a fake group per scenario/method; for video, the source domain has only real and \texttt{Open-Sora}, and target groups are defined per generator.

For two groups with parameters $(\mu_1,\Sigma_1)$ and $(\mu_2,\Sigma_2)$, the Fr\'echet distance is
\begin{equation}
    d_F\!\big((\mu_1,\Sigma_1),(\mu_2,\Sigma_2)\big) = \|\mu_1-\mu_2\|_2^2 + \mathrm{tr}(\Sigma_1) + \mathrm{tr}(\Sigma_2) - 2\,\mathrm{tr}\!\left[(\Sigma_1\Sigma_2)^{1/2}\right].
\end{equation}
For each layer, we compute the full source-by-target distance matrix and derive three summaries per target group $g$:
\begin{equation}
    d_{\text{real}}(g,\ell), \qquad
    d_{\text{spoof}}(g,\ell) = \min_{h \in \mathcal{S}_{\text{train}}} d_F\!\big((\mu_h^{(\ell)},\Sigma_h^{(\ell)}),(\mu_g^{(\ell)},\Sigma_g^{(\ell)})\big), \qquad
    \Delta(g,\ell) = d_{\text{real}}(g,\ell) - d_{\text{spoof}}(g,\ell),
\end{equation}
where $\mathcal{S}_{\text{train}}$ is the set of source-domain spoof groups. Intuitively, $\Delta$ measures whether a target group lies closer to the source real region or to the source fake region. We test whether geometry explains performance by correlating each measure with per-generator probe performance using both Pearson $r$ and Spearman $\rho$, computed independently per layer and per probe. We treat $\Delta$ as a \emph{relative-position diagnostic}, not as a causal account of detection.

\section{Results}\label{sec:results}

We organize the results around the two claims of the audit. Section~\ref{sec:probe-results} reports linear-probe performance across benchmarks, showing that frozen-SSL representations expose substantial real-versus-fake structure and approach reported specialized detectors. Section~\ref{sec:geometry-results} then asks why some target generators are harder than others, and shows that the relative Fr\'echet margin $\Delta$ is the most consistent geometric predictor of difficulty across all three modalities.

\subsection{Linear probes reveal strong benchmark separability}\label{sec:probe-results}

\subsubsection{Video: AIGVDBench}\label{sec:results-video}

\begin{figure}[!ht]
    \centering
    \begin{subfigure}[b]{0.65\textwidth}
        \centering
        \includegraphics[width=\textwidth]{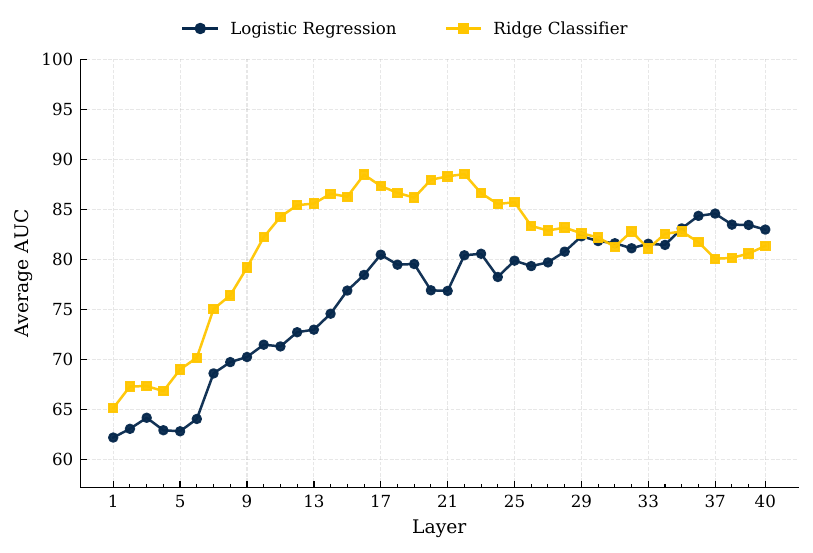}
    \end{subfigure}
    \hfill
    \begin{subfigure}[b]{0.30\textwidth}
        \vspace{0pt}
        \centering
        \raisebox{3.3cm}{%
        \begin{minipage}{\textwidth}
            \centering
            \small
            \setlength{\tabcolsep}{5pt}
            \begin{tabular}{lc}
                \toprule
                \textbf{Detector} & \textbf{AUC} \\
                \midrule
                \multicolumn{2}{l}{\textit{AIGVDBench top-10}} \\
                \midrule
                X3D~\cite{feichtenhofer2020x3d}              & 77.67 \\
                UnivFD~\cite{ojha2023towards}                & 78.44 \\
                DeCoF~\cite{ma2025detecting}                 & 79.13 \\
                CNNSpot~\cite{wang2020cnn}                   & 80.28 \\
                D3~\cite{zheng2025d3}                        & 80.39 \\
                TimeSformer~\cite{bertasius2021space}         & 83.27 \\
                UniFormerV2~\cite{li2023uniformerv2}          & 84.62 \\
                ForgeLens3~\cite{chen2025forgelens}           & 85.08 \\
                ForgeLens1~\cite{chen2025forgelens}           & 89.41 \\
                \textbf{Effort}~\cite{yan2024effort}          & \textbf{89.82} \\
                \midrule
                \multicolumn{2}{l}{\textit{Our top LR/Ridge probes}} \\
                \midrule
                LR L37                                        & 84.56 \\
                \textbf{Ridge L22}                            & \textbf{88.51} \\
                \bottomrule
            \end{tabular}
        \end{minipage}}
    \end{subfigure}
    \caption{\small AIGVDBench. (a) Macro-average per-generator AUC on the full test split as a function of V-JEPA2 layer for logistic regression (LR) and ridge probes. (b) Reported overall average AUC across 20 open-source and 11 closed-source generators for the top-10 detectors in the AIGVDBench paper and our top LR/Ridge probes.}
    \label{fig:aigvdbench}
\end{figure}

The strongest frozen-SSL probe on AIGVDBench is the layer-22 ridge classifier, which reaches \textbf{88.51} macro-average per-generator AUC (Figure~\ref{fig:aigvdbench}a). This is close to the strongest reported detectors in the benchmark paper, ForgeLens1 (89.41) and Effort (89.82), and exceeds every other detector listed (Figure~\ref{fig:aigvdbench}b). The probe uses no benchmark-specific architecture, no fine-tuning, and no temporal reasoning module beyond what V-JEPA2 itself provides; it is a single linear classifier on layer-pooled representations.

Layer depth matters. At layer 1, logistic regression and ridge start at 62.16 and 65.10 AUC respectively. Ridge improves rapidly and peaks at 88.51 at layer 22, while logistic regression continues to improve more gradually and peaks later, at layer 37, with 84.56 AUC. Real-versus-fake structure is therefore not uniformly distributed across the V-JEPA2 hierarchy, but becomes much more linearly accessible in higher-level representations.

The aggregate is not driven by a few easy generators (Figure~\ref{fig:aigvdbench-pergen}). At the layer-22 ridge configuration, the median per-generator AUC is 92.50 and the range is 63.00 to 100.00. The highest values are \texttt{Open-Sora} (100.0, the source generator), \texttt{SEINE} (99.98), and \texttt{Pyramid-Flow} (99.98); the lowest are \texttt{wan} (63.00), \texttt{pika} (66.00), and \texttt{Luma} (69.60). The full per-generator breakdown is shown in Appendix Figure~\ref{fig:aigvdbench-pergen}.

\subsubsection{Image: Celeb-DF++}\label{sec:results-image}

\begin{figure}[!ht]
    \centering
    \begin{subfigure}[b]{0.65\textwidth}
        \centering
        \includegraphics[width=\textwidth]{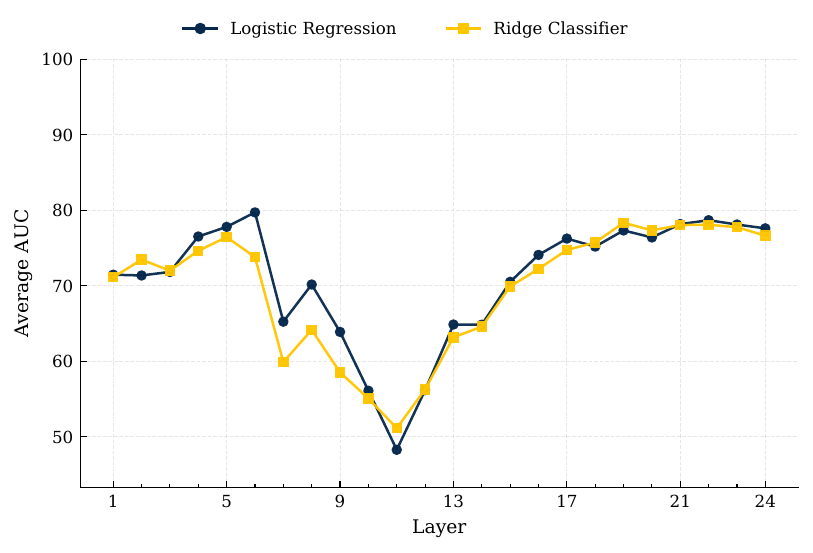}
    \end{subfigure}
    \hfill
    \begin{subfigure}[b]{0.30\textwidth}
        \vspace{0pt}
        \centering
        \raisebox{3.2cm}{%
        \begin{minipage}{\textwidth}
            \centering
            \small
            \setlength{\tabcolsep}{5pt}
            \begin{tabular}{lc}
                \toprule
                \textbf{Detector} & \textbf{AUC} \\
                \midrule
                \multicolumn{2}{l}{\textit{Celeb-DF++ reported detectors}} \\
                \midrule
                Xception~\cite{chollet2017xception}          & 72.3 \\
                RFM~\cite{wang2020cnn}                       & 71.0 \\
                CLIP~\cite{radford2021learning}              & 69.1 \\
                SIA~\cite{sun2022information}                & 70.3 \\
                UCF~\cite{yan2023ucf}                        & 67.5 \\
                IID~\cite{huang2023implicit}                 & 68.7 \\
                ProDet~\cite{cheng2024can}                   & 71.4 \\
                \textbf{Effort}~\cite{yan2024effort}         & \textbf{83.0} \\
                \midrule
                \multicolumn{2}{l}{\textit{Our top LR/Ridge probes}} \\
                \midrule
                \textbf{LR L6}                               & \textbf{79.72} \\
                Ridge L19                                    & 78.34 \\
                \bottomrule
            \end{tabular}
        \end{minipage}}
    \end{subfigure}
    \caption{\small Celeb-DF++. (a) Layer-wise macro-average per-generator AUC on the Celeb-DF++ GF-eval test split for logistic regression (LR) and ridge probes on DINOv3 representations. (b) Reported frame-level average AUC from Celeb-DF++ paper for detectors trained on Celeb-DF and evaluated on Celeb-DF++; and our top LR/Ridge probes.}
    \label{fig:celebdf}
\end{figure}

On Celeb-DF++ (Figure~\ref{fig:celebdf}a), the strongest frozen-SSL probe is logistic regression at layer 6, reaching \textbf{79.72} AUC; ridge peaks later, at layer 19, with 78.34. This trails the strongest reported detector, Effort (83.0), but is competitive with or above the other listed baselines (Xception 72.3, RFM 71.0, CLIP 69.1, SIA 70.3, UCF 67.5, IID 68.7, ProDet 71.4).

The layer profile is markedly non-monotonic, in contrast with the AIGVDBench setting. Both probes start in the low 70s, rise sharply in the shallow layers, collapse in the middle of the network, and recover later. Logistic regression drops from 79.72 at layer 6 to 48.29 at layer 11; ridge drops from 76.45 at layer 5 to 51.14 at layer 11. The most useful DINOv3 representations for this benchmark are therefore not the deepest ones; separability varies substantially across the SSL hierarchy.

At the best configuration the median per-generator AUC is 82.22, slightly above the 79.72 macro-average, and per-generator AUC ranges from 60.45 to 94.81 (Figure~\ref{fig:celebdf-pergen}). The benchmark is broadly separable but with substantial diversity across the FS/FR/TF scenarios. The full per-generator breakdown is shown in Appendix Figure~\ref{fig:celebdf-pergen}.

\subsubsection{Audio: ASVspoof2019 LA and English MLAAD}\label{sec:results-audio}

The audio setting requires a slightly different framing. The original \emph{MLAAD} evaluation numbers are reported on MLAAD~v1, which is now two years old and substantially smaller than MLAAD~v9, and our evaluation is restricted to English (since ASVspoof2019 LA is English-only). A like-for-like comparison with the original MLAAD numbers is therefore not available. We instead validate the probe on ASVspoof2019 LA evaluation protocol as a sanity check, then transfer to the MLAAD~v9 English split.

\Paragraph{ASVspoof2019 LA evaluation protocol sanity check.} The best logistic-regression probe (layer~6) reaches \textbf{4.78\% EER}, and the best ridge probe (layer~11) reaches 5.70\% EER. The top reported single systems on ASVspoof2019 LA are T24 (4.04), T45 (5.06), T04 (5.74), T01 (5.97), and T39 (7.01)~\cite{nautsch2021asvspoof}. The layer-6 logistic-regression probe is worse than only the strongest reported single system and outperforms the other four listed systems; the best ridge probe falls within the same range. This confirms that frozen XLS-R representations paired with a linear probe are competitive on the canonical source benchmark, not just on transfer.

\begin{figure}[!ht]
    \centering
    \begin{subfigure}[b]{0.65\textwidth}
        \centering
        \includegraphics[width=\textwidth]{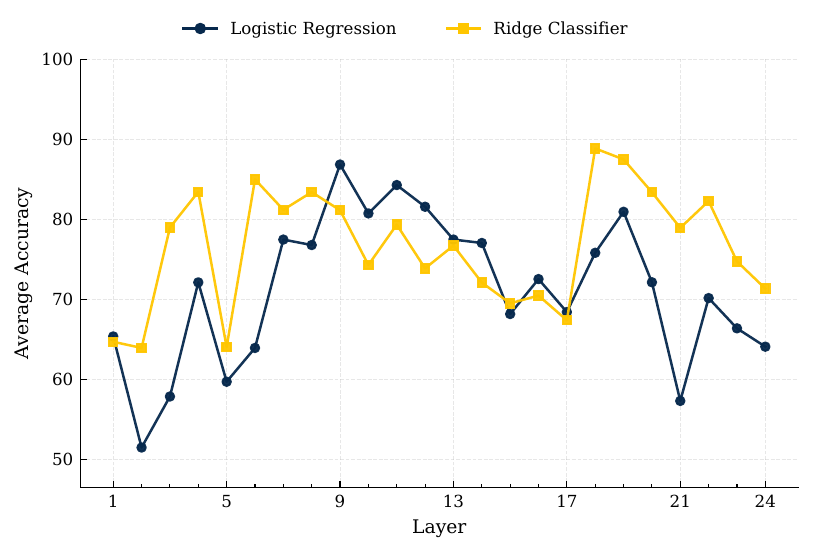}
    \end{subfigure}
    \hfill
    \begin{subfigure}[b]{0.30\textwidth}
        \vspace{0pt}
        \centering
        \raisebox{3cm}{%
        \begin{minipage}{\textwidth}
            \centering
            \small
            \setlength{\tabcolsep}{5pt}
            \begin{tabular}{lc}
                \toprule
                \textbf{Detector} & \textbf{EER} \\
                \midrule
                \multicolumn{2}{l}{\textit{ASVspoof2019 LA top-5}} \\
                \midrule
                T39            & 7.01 \\
                T01            & 5.97 \\
                T04            & 5.74 \\
                T45            & 5.06 \\
                \textbf{T24}   & \textbf{4.04} \\
                \midrule
                \multicolumn{2}{l}{\textit{Our top-5 probes (LA)}} \\
                \midrule
                Ridge L11      & 5.70 \\
                LR L19         & 5.28 \\
                LR L8          & 5.19 \\
                LR L7          & 4.92 \\
                \textbf{LR L6} & \textbf{4.78} \\
                \bottomrule
            \end{tabular}
        \end{minipage}}
    \end{subfigure}
    \caption{\small Audio. (a) Layer-wise macro-average per-generator spoof accuracy on the MLAAD~v9 English split (84 generators) for logistic regression (LR) and ridge probes on XLS-R representations. (b) Top-5 single systems on ASVspoof2019 LA by EER, alongside the top-5 of our probes on the same LA evaluation split.}
    \label{fig:mlaad}
\end{figure}

\Paragraph{Transfer to English MLAAD.} On the current English MLAAD split (84 target generators), logistic regression peaks at layer~9 with 86.83\% mean per-generator accuracy, and ridge peaks at layer~18 with \textbf{88.84\%}, the best overall audio-transfer result (Figure~\ref{fig:mlaad}a). The layer profile is non-monotonic: ridge dominates through much of the network, especially in shallow and late-middle layers, while logistic regression is most competitive around layers~9--14.

At the layer-18 ridge configuration (Figure~\ref{fig:mlaad-pergen}), the median per-generator accuracy is 93.1\% and the range is 39.2\% to 100.0\%. Several systems are detected perfectly, including \texttt{WhisperSpeech}, \texttt{tts\_models\_en\_sam\_tacotron-DDC}, \texttt{e2-tts}, \texttt{Indri-TTS-0.1}, and \texttt{facebook\_mms-tts-eng}. Harder systems include \texttt{DeepGram} (39.2\%), \texttt{Kyutai-TTS} (48.2\%), and \texttt{Microsoft VibeVoice 1.5B} (52.0\%). The full per-generator breakdown is shown in Appendix Figure~\ref{fig:mlaad-pergen}.

\Paragraph{Takeaway.} Across all evaluated settings, frozen modality-specific SSL representations paired with a linear probe recover most of the reported detector signal. Specifically, within $\sim$1.3 AUC of the strongest reported detector on video (exceeding eight of the top-10), between the strongest detector and the rest of the leaderboard on image, and competitive with top single systems on ASVspoof2019 LA while reaching $88.84\%$ spoof-only generator accuracy on MLAAD~v9 English. None of these probes uses task-specific architecture or fine-tuning beyond what the SSL backbone provides. These benchmarks already contain substantial linearly accessible signal in generic pretrained representation spaces. We next ask why generator-level difficulty varies the way it does.

\subsection{Feature-space geometry}\label{sec:geometry-results}

We now correlate per-generator probe performance with the three Fr\'echet-distance summaries from Section~\ref{sec:method}: $d_{\text{real}}$, $d_{\text{spoof}}$, and $\Delta = d_{\text{real}} - d_{\text{spoof}}$. The analysis is performed independently at each SSL layer; we report Pearson $r$ and Spearman $\rho$.

\subsubsection{Video: AIGVDBench}

\begin{figure}[!ht]
    \centering
    \includegraphics[width=\textwidth]{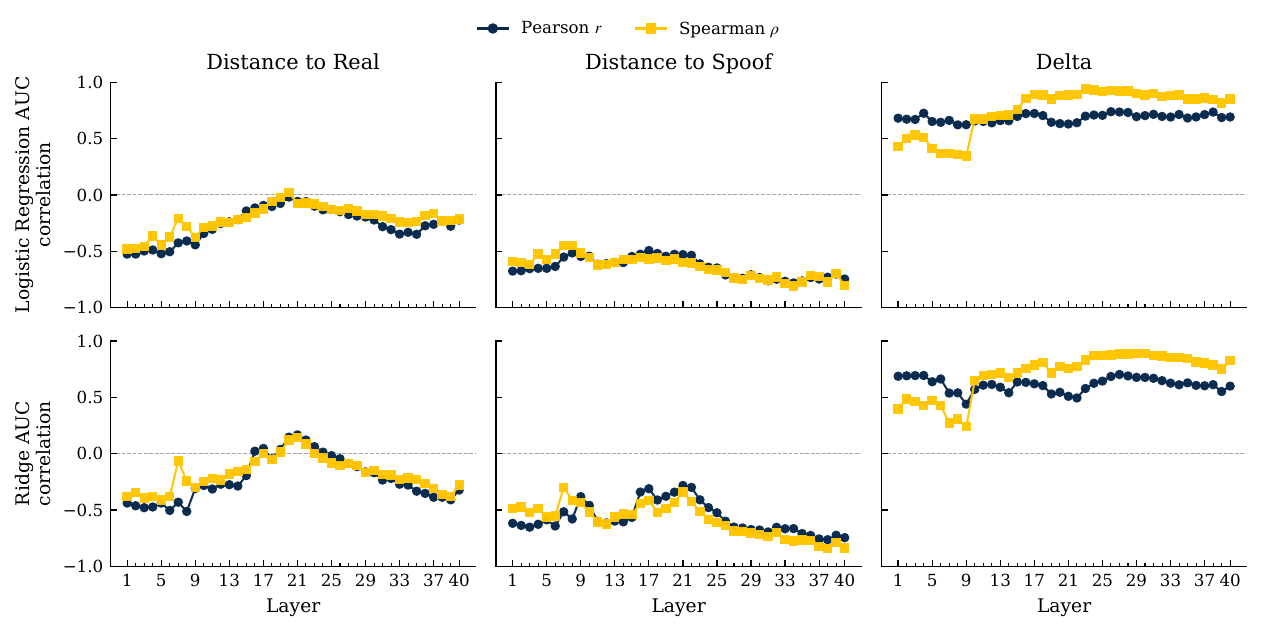}
    \caption{\small AIGVDBench: layer-wise Pearson and Spearman correlations between per-generator AUC and the three Fr\'echet-distance summaries $d_{\text{real}}$, $d_{\text{spoof}}$, $\Delta$ for logistic regression (LR) and ridge probes on V-JEPA2 representations.}
    \label{fig:aigvdbench-corr}
\end{figure}

Across most layers, $d_{\text{spoof}}$ is negatively correlated with performance and $\Delta$ is positively correlated with performance, while $d_{\text{real}}$ is weaker and less stable (Figure~\ref{fig:aigvdbench-corr}). Generators that lie closer to the source fake region (low $d_{\text{spoof}}$, high $\Delta$) tend to be easier to detect. The strongest $\Delta$ correlations reach approximately $r \approx 0.74$ and $\rho \approx 0.94$ for logistic regression, and $r \approx 0.70$ and $\rho \approx 0.89$ for ridge. These are the most consistent positive associations in the figure. The strongest geometry--performance relationships occur in the same middle-to-late region of the network where the layer-22 ridge probe reaches its peak.

\subsubsection{Image: Celeb-DF++}

\begin{figure}[!ht]
    \centering
    \includegraphics[width=\textwidth]{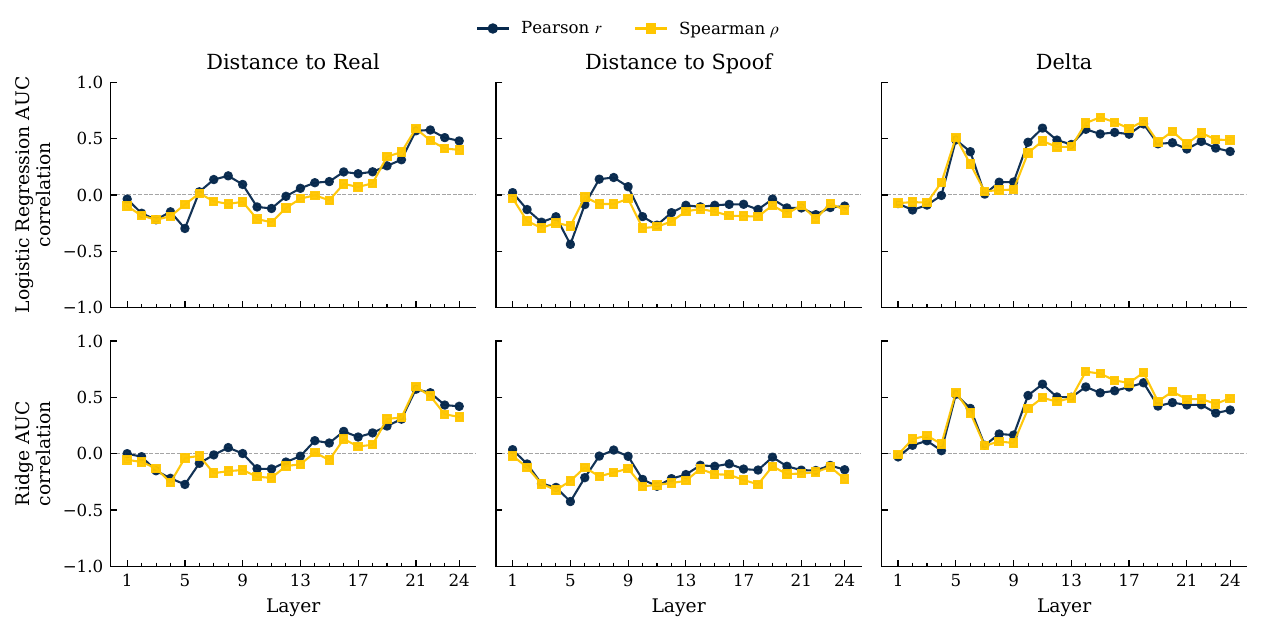}
    \caption{\small Celeb-DF++: layer-wise Pearson and Spearman correlations between per-generator AUC and the three Fr\'echet-distance summaries for logistic regression (LR) and ridge probes on DINOv3 representations.}
    \label{fig:celebdf-corr}
\end{figure}

On Celeb-DF++ (Figure~\ref{fig:celebdf-corr}), $\Delta$ again provides the strongest overall explanation, but the pattern is more nuanced than in the video setting. For logistic regression, the strongest Pearson correlation is $r = 0.630$ at layer~18 and the strongest Spearman is $\rho = 0.689$ at layer~15; for ridge, $r = 0.629$ at layer~18 and $\rho = 0.728$ at layer~14. The absolute distance to source real is informative but weaker, peaking at $r = 0.576$, $\rho = 0.589$ for logistic regression and $r = 0.572$, $\rho = 0.593$ for ridge. The strongest negative $d_{\text{spoof}}$ correlations are weaker still ($\sim$$-0.43$ for both probes).

A noteworthy observation: at the best probe layer (layer~6, logistic regression), $\Delta$ is only moderately associated with performance ($r = 0.384$, $\rho = 0.274$ for LR; $r = 0.402$, $\rho = 0.362$ for ridge). The strongest-performing layer is therefore not the most geometrically interpretable layer; the recovery phase later in the DINOv3 hierarchy is where generator position becomes most predictive of detection difficulty. This dissociation suggests that, in the image setting, shallow layers can support strong probe performance for reasons not fully captured by second-order distributional summaries.

\subsubsection{Audio: English MLAAD}

\begin{figure}[!ht]
    \centering
    \includegraphics[width=\textwidth]{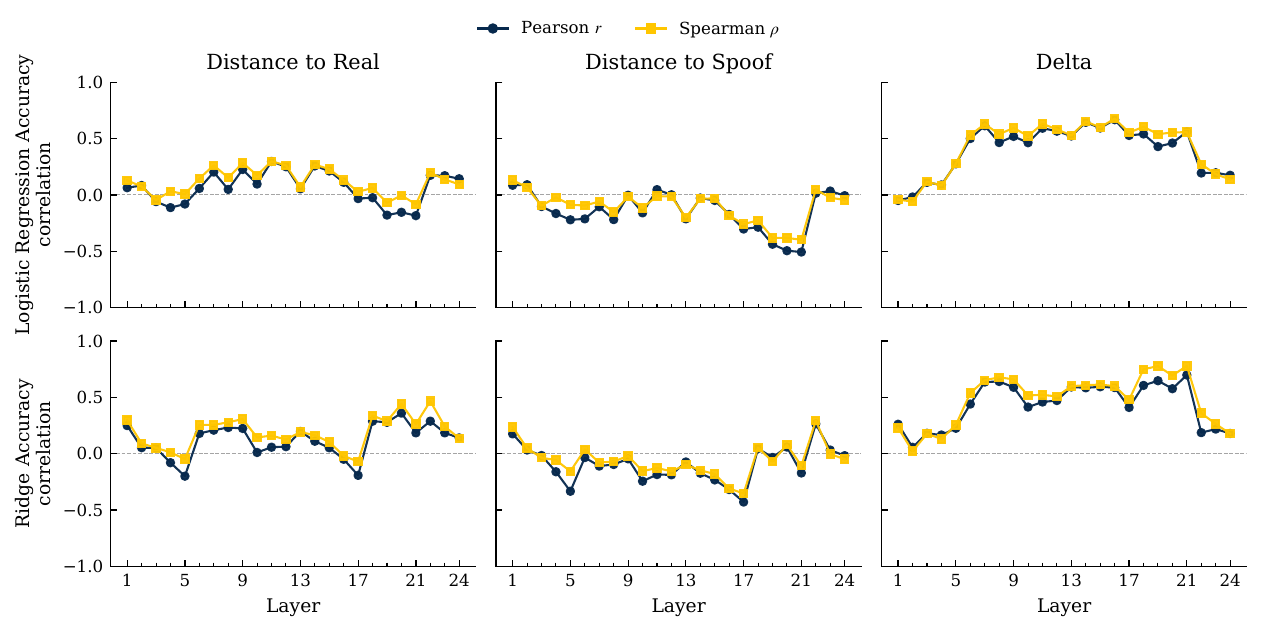}
    \caption{\small English MLAAD: layer-wise Pearson and Spearman correlations between per-generator spoof accuracy and the three Fr\'echet-distance summaries for logistic regression (LR) and ridge probes on XLS-R representations.}
    \label{fig:mlaad-corr}
\end{figure}

On MLAAD (Figure~\ref{fig:mlaad-corr}), $\Delta$ again gives the strongest geometry--performance relationship. For logistic regression, $\Delta$ peaks at layer~16 with $r = 0.668$ and $\rho = 0.678$; for ridge, $\Delta$ peaks at layer~21 with $r = 0.698$ and $\rho = 0.779$. These are the strongest associations observed in the audio setting. The absolute distance to source real is much weaker (peaking at $r = 0.297$, $\rho = 0.298$ for LR; $r = 0.360$, $\rho = 0.466$ for ridge), and the strongest negative $d_{\text{spoof}}$ correlations are intermediate ($r = -0.507$, $\rho = -0.398$ for LR; $r = -0.430$, $\rho = -0.354$ for ridge).

The layers with the strongest geometry--performance relationships also align with the strongest probe region: the best probe is layer-18 ridge, and the strongest $\Delta$ correlations occur at layers~16 (LR) and~21 (ridge). As in the video setting, the upper-level XLS-R representations that make the benchmark most linearly separable are also the ones in which generator position best explains detection difficulty.

\Paragraph{Takeaway.} Across all three modalities, the most consistent geometry signal is the relative margin $\Delta = d_{\text{real}} - d_{\text{spoof}}$, not the absolute distance to either region. Target generators are easier when they lie closer to source fake structure and harder when they lie closer to source real structure. This carries a structural implication for benchmark composition: target generators with large positive $\Delta$ recapture structure already present in the source spoof distribution, and their contribution to aggregate scores is partly redundant with generators the probe has effectively already seen. The image benchmark is a partial exception: its best probe performance occurs in shallow DINOv3 layers, while $\Delta$ explains generator-level difficulty most strongly in later layers. This suggests that feature-space geometry should be interpreted layer-wise rather than reduced to a single benchmark-level score.

\section{Discussion}\label{sec:discussion}

\Paragraph{Limitations.}
The geometry should not be overclaimed. It is a diagnostic, not a complete causal theory of detection. It uses the same feature space as the probes; it models generator groups as Gaussians; and it summarizes each group with second-order statistics. Some generators may be easy or hard for reasons not captured by these summaries. The audio result also requires careful framing: ASVspoof2019 LA validates that the XLS-R probe is competitive on a canonical source benchmark, but the MLAAD evaluation is spoof-only, which means our MLAAD number is best described as generator-level spoof recall rather than full target-domain real-versus-fake detection. We restrict MLAAD to English to isolate generator shift from cross-lingual mismatch; multilingual MLAAD evaluation would require a multilingual source benchmark and is outside the scope of this audit. Finally, our audit covers one contemporary, generator-diverse benchmark per modality. Whether the same patterns generalize to other deepfake benchmarks, we leave to future work.

\Paragraph{Summary and broader impact.}
The central result is not that linear probes are the best deepfake detectors; it is that a surprising amount of benchmark performance is already accessible from frozen general-purpose SSL representations and a linear classifier. 

The SSLs we use were trained on large amounts of unlabeled natural data for general modality understanding, with no deepfake supervision and no exposure to generator outputs as a distinct class. The fact that real-versus-fake is nonetheless linearly accessible in their feature spaces means the benchmarks' fakes are separable from reals along directions a representation of natural data already encodes for other reasons.

This supports the two implications raised earlier. First, a benchmark whose labels are linearly accessible from a generic representation is rewarding general modality understanding more than forensic understanding, and is unlikely to reflect realistic threat models in which adversaries continue to iterate generators and post-processing toward natural data. Second, when bespoke detectors only marginally exceed such a probe on aggregate scores, the gap does not establish that they have learned forensic structure; the burden of evidence shifts to showing performance precisely where generic separability fails.

The practical recommendation that follows is that a frozen-SSL linear probe should be a standard sanity check during benchmark construction: a benchmark that a probe can already solve to some degree has, by construction, real and fake samples that are not entangled in general-purpose representation space, and strong scores on it should be read as evidence about the benchmark distribution as much as about the detector. The geometry analysis plays a supporting role here, not a prescriptive one. We do not recommend $\Delta$ as a design target, because an ideal benchmark would be one in which $\Delta$ is uninformative because reals and fakes are not cleanly separable to begin with. Its value in this paper is diagnostic: it explains why some generators are easier than others for the probe, and it surfaces a structural concern about current test sets, where some generators are effectively redundant because they sit close to already-seen spoof structure rather than testing genuinely novel generators.

\newpage
\bibliographystyle{plain}
\bibliography{refs}


\clearpage
\appendix

\section{Appendix}

\begin{figure}[!ht]
    \centering
    \includegraphics[width=1.00\textwidth]{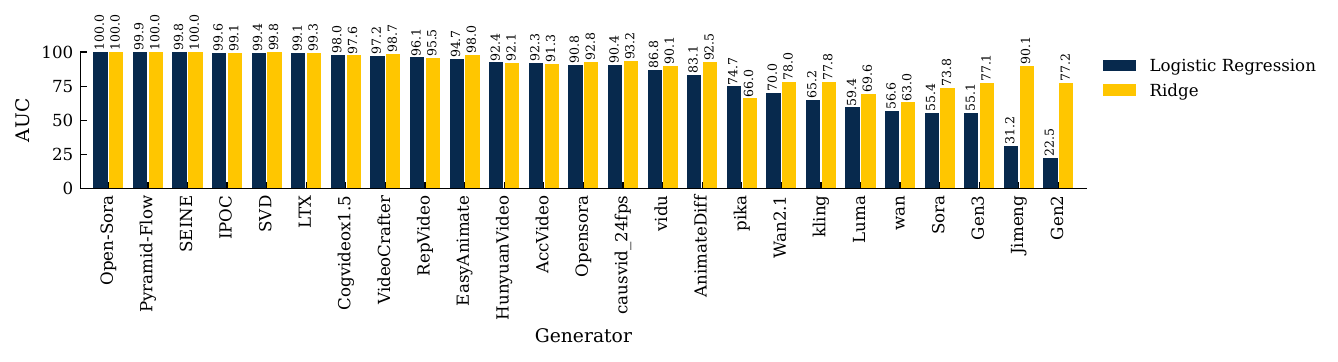}
    \caption{\small AIGVDBench: per-generator AUC at the strongest probe configuration (V-JEPA2 layer 22, ridge). Macro-average 88.51, median 92.50, range 63.00--100.00. Strong performance is broadly distributed; a small number of generators (\texttt{wan}, \texttt{pika}, \texttt{Luma}) remain meaningfully harder. \texttt{Open-Sora} and \texttt{Opensora} are distinct AIGVDBench generator labels: \texttt{Open-Sora} denotes the open-source generator used for source training, while \texttt{Opensora} denotes the separately listed generator from the benchmark's closed-source directory.}
    \label{fig:aigvdbench-pergen}
\end{figure}

\begin{figure}[!ht]
    \centering
    \includegraphics[width=1.00\textwidth]{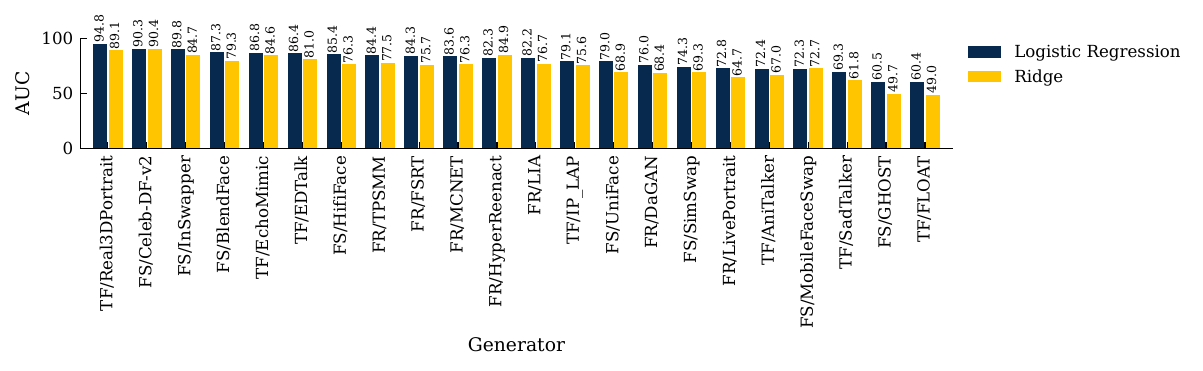}
    \caption{\small Celeb-DF++: per-generator AUC at the strongest probe configuration (DINOv3 layer 6, logistic regression). Macro-average 79.72, median 82.22, range 60.45--94.81 across face-swap, face-reenactment, and talking-face methods.}
    \label{fig:celebdf-pergen}
\end{figure}

\appendix

\paragraph{Compute resources.} All experiments used frozen pretrained backbones; we did not fine-tune V-JEPA2, DINOv3, or XLS-R. The main computational cost was one-time feature extraction, followed by substantially cheaper linear probing and representation-space geometry analysis. Feature extraction was performed on a single NVIDIA A100 GPU with 80GB of VRAM. The extraction pipeline processed inputs in batches, saved pooled hidden-layer representations to chunked Parquet files, and supported resuming from already written chunks. The probing and geometry analyses were then run on the saved representations and did not require GPU training.

The raw datasets, downloaded archives, extracted files, cached pretrained model weights, intermediate files, and processed layer-wise representations require at least 1TB of disk storage. Storage is a significant practical requirement because the pipeline saves representations for each example and each hidden layer before running the linear probes and Fréchet-geometry analysis.

We did not keep exact wall-clock logs for all runs, so we report only the hardware class and storage requirement rather than precise runtimes. In our runs, feature extraction was the dominant cost, especially for the video benchmark, while fitting the linear probes and computing generator-level geometry summaries were comparatively lightweight once the representations had been written to disk.


\begin{figure}[!ht]
    \centering
    \includegraphics[width=\textwidth]{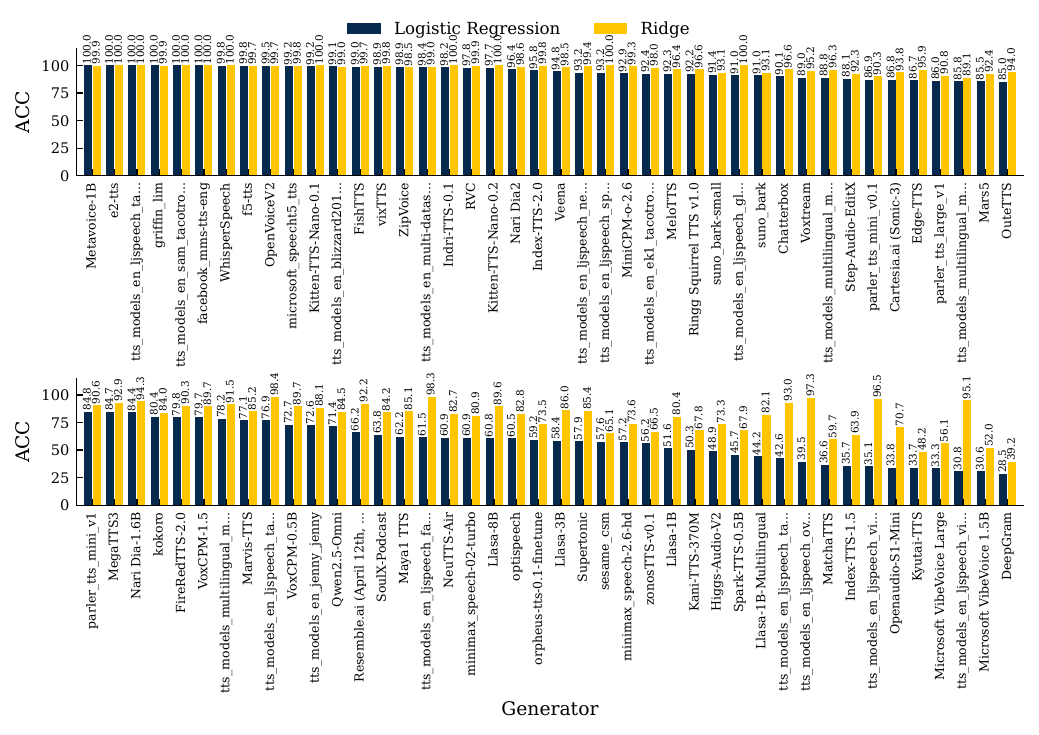}
    \caption{\small English MLAAD: per-generator spoof accuracy at the strongest probe configuration (XLS-R layer 18, ridge). Macro-average 88.84\%, median 93.1\%, range 39.2--100.0\% across 84 TTS systems.}
    \label{fig:mlaad-pergen}
\end{figure}
\clearpage


\end{document}